\documentclass[11pt]{article}

\usepackage[preprint]{acl}

\usepackage{times}
\usepackage{latexsym}

\usepackage[T1]{fontenc}
\usepackage[utf8]{inputenc}

\usepackage{microtype}

\usepackage{inconsolata}

\usepackage{graphicx}

\usepackage{array}
\usepackage{multirow}
\usepackage{xspace}
\usepackage{enumerate}
\usepackage{pifont}
\usepackage{paralist}
\usepackage{amsmath}
\usepackage{amssymb}
\usepackage{subcaption}
\usepackage{xcolor}
\usepackage{colortbl}
\usepackage[normalem]{ulem}
\usepackage{booktabs}
\usepackage{bbm}
\usepackage{CJKutf8}
\usepackage[most]{tcolorbox}

\usepackage{hyperref}

\newcommand{\githubicon}{\raisebox{-0.15ex}{\includegraphics[height=1.8ex]{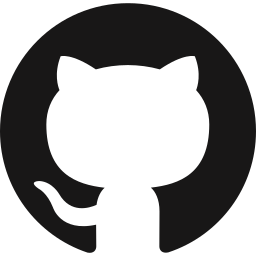}}}
\newcommand{\huggingfaceicon}{\raisebox{-0.25ex}{\includegraphics[height=2ex]{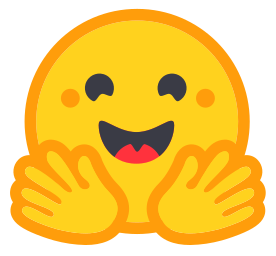}}}

\title{MultiModal Code-Switching: Interleaving Visual Objects into Language for Explicit Object-Level Alignment}

\author{
    Changhao Xiang \quad Shangyu Xing \quad Zhen Wu{\thanks{~~~Corresponding author.}} \quad Jianbing Zhang \quad Xinyu Dai \\
    National Key Laboratory for Novel Software Technology, Nanjing University, China \\
    {\tt \{xiangch, xsy\}@smail.nju.edu.cn} \quad
    {\tt \{wuz, zjb, daixinyu\}@nju.edu.cn} \\
    \href{https://github.com/Changhao-Xiang/MM-CodeSwitch}{\githubicon\ \textbf{Code}}
    \quad
    \href{https://huggingface.co/datasets/LockOnN/MMCS-Data}{\huggingfaceicon\ \textbf{Dataset}}
}

\begin{document}
\maketitle
\begin{abstract}
Existing Multimodal Large Language Models (MLLMs) predominantly rely on image-text pairs for modality alignment pretraining, mapping global image representations to long textual descriptions. 
However, this image-level alignment suffers from \textit{referential ambiguity}: models struggle to infer the correspondences between multiple visual objects and textual entities from the global representation, leading to data inefficiency and suboptimal semantic grounding. 
To address this, we propose MultiModal Code-Switching (MMCS), a novel pretraining paradigm that provides explicit object-level supervision. Inspired by the linguistic phenomenon of code-switching, MMCS interleaves vision and language by replacing textual entities with their corresponding visual objects, enforcing local vision-language grounding.
We further develop a scalable data synthesis pipeline to generate a pretraining dataset of 773K samples with accurate object–entity correspondences. Experiments show that MMCS is highly data-efficient: with only 50K samples, it matches or surpasses models trained on 600K image–text pairs. Furthermore, MMCS consistently improves visual grounding and perception capabilities across varying model scales.

\end{abstract}

\section{Introduction}
\label{sec:intro}

\begin{figure*}[t]
    \centering
    \begin{minipage}[t]{0.6\textwidth}
        \centering
        \includegraphics[width=1.0\linewidth]{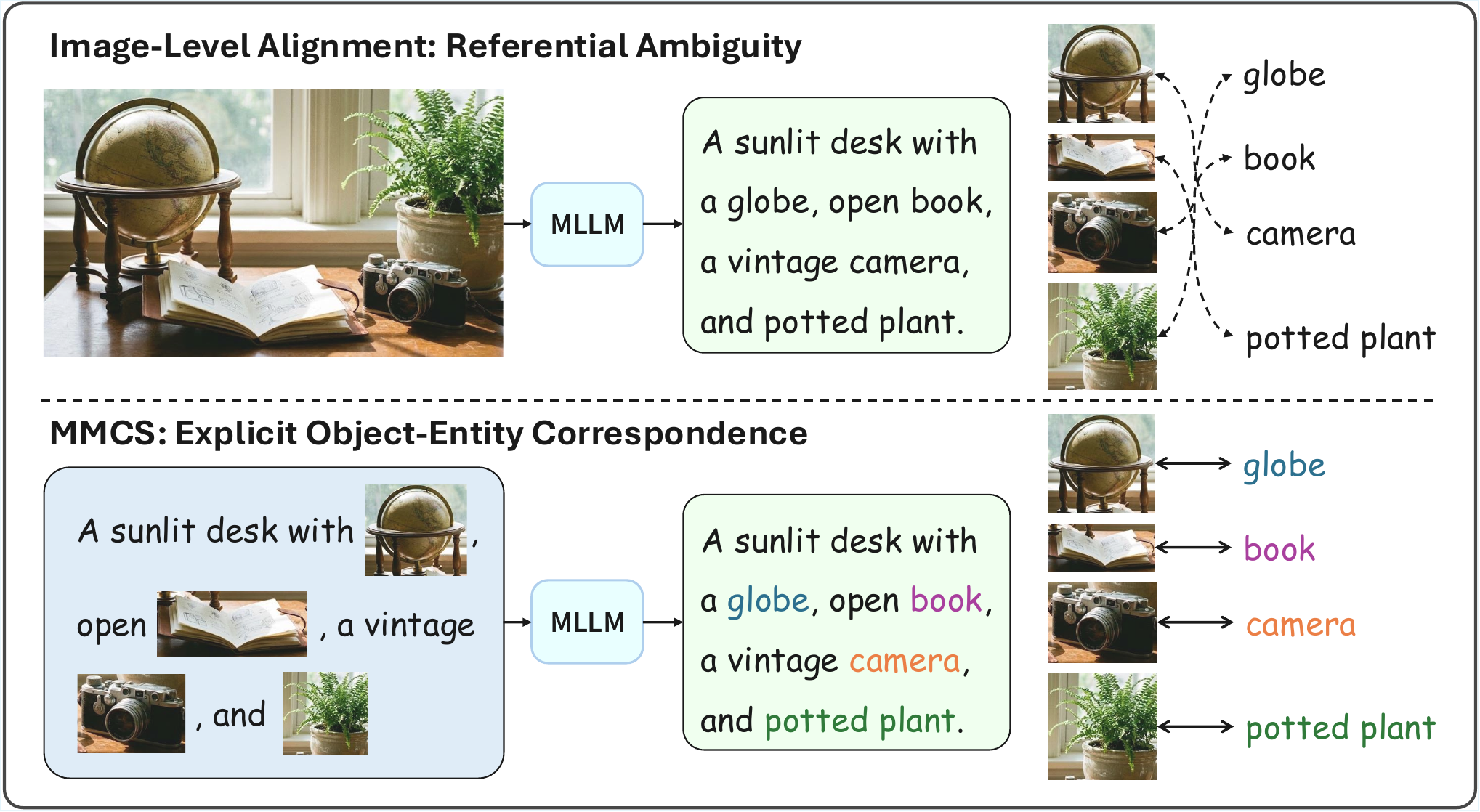}
    \end{minipage}\hfill
    \begin{minipage}[t]{0.38\linewidth}
        \centering
        \includegraphics[width=0.95\linewidth]{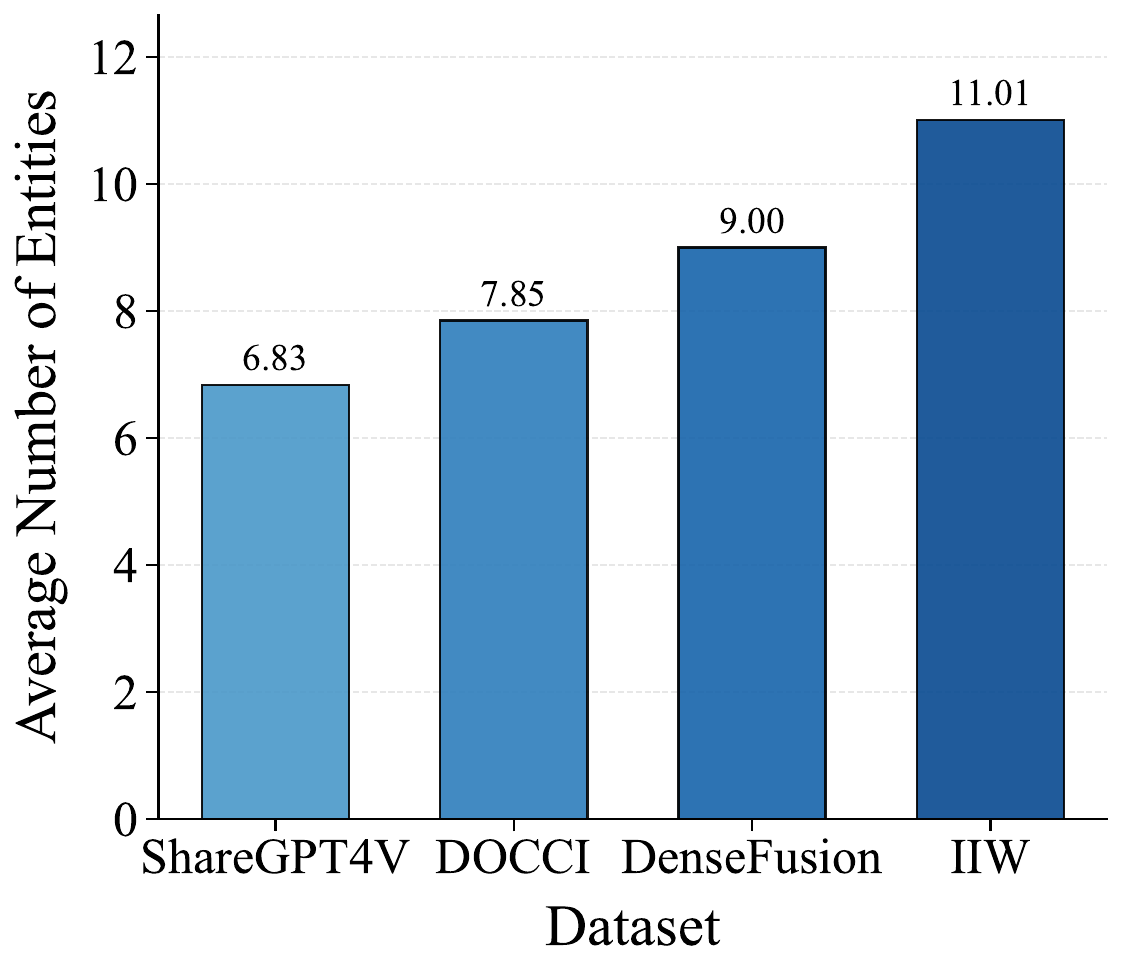}
    \end{minipage}
    \caption{\textbf{Left}: Illustration of the referential ambiguity in standard image-level alignment (top), contrasted with our MultiModal Code-Switching (MMCS) paradigm (bottom). MMCS resolves ambiguity by replacing textual entities with their corresponding visual objects to provide explicit correspondence signals. \textbf{Right:} The large average number of entities in dense caption datasets highlights the complexity of natural images.}
    \label{fig:ambiguity_of_image_caption}
\end{figure*}

Multimodal Large Language Models (MLLMs) have established a new state-of-the-art in vision-language understanding, demonstrating exceptional performance across visual question answering \citep{vqav2,gqa}, document understanding \citep{docvqa}, and visual grounding \citep{refcoco,refcocog}. The dominant architecture for these models \citep{flamingo,llava,instructblip,qwen25vl,internvl3} generally comprises a vision encoder, a Large Language Model (LLM) backbone, and a projector (e.g., MLP or Q-Former) that bridges the modality gap. To unify these components, the standard training paradigm follows a two-stage process: 1) modality alignment pretraining, mapping visual features into the LLM’s semantic space via image-text pairs; and 2) multimodal supervised instruction tuning (SFT), optimizing the model for downstream task execution. Within this framework, modality alignment is foundational, as the fidelity of this alignment dictates the upper bound of the model's multimodal capabilities \citep{llava,mm1}. 

The prevailing consensus in recent research is to utilize dense image captions for modality alignment, providing detail-rich linguistic signals  \citep{allava,sharegpt4v,densefusion,denseworld,molmo,caprl}. In this paradigm, the model encodes the image into a global image representation, and is then trained to predict a lengthy text sequence based on this global representation, thereby achieving alignment at the image level. However, natural images are inherently complex, often containing multiple objects and background elements. As highlighted in Figure \ref{fig:ambiguity_of_image_caption} (right), standard dense caption datasets contain an average of 6.8 to 11.0 distinct entities per sample \citep{sharegpt4v,docci,densefusion,imageinwords}. While the captions meticulously enumerate specific objects and attributes, the vision encoder and projector compress the entire scene into a generic, global representation.

This discrepancy leads to the issue of \textit{referential ambiguity}: the model must implicitly infer the correspondence between specific visual regions and the corresponding textual phrases from global representations. From a computational perspective, this ``many-to-many'' mapping forces the model to rely on statistical co-occurrences rather than genuine semantic grounding. Consequently, this implicit alignment paradigm is highly data-inefficient, necessitating massive-scale datasets to learn robust object-entity associations \citep{mm1,sail_vl}. Our analysis substantiates this by revealing that models pretrained with standard image-caption pairs exhibit diffuse attention patterns and suboptimal representation consistency.

To address these limitations, we propose MultiModal Code-Switching (MMCS), a novel pretraining paradigm that introduces \textit{explicit object-level supervision}. Inspired by the linguistic phenomenon of code-switching \citep{cs,cs_2018}, MMCS treats vision and language as distinct ``codes''. Instead of relying solely on global image context, we create interleaved representations by substituting the embeddings of textual entities with the embeddings of their corresponding visual objects (Figure \ref{fig:ambiguity_of_image_caption}, left). By conditioning the generation of the immediate textual context directly on these local visual features, MMCS imposes a structural constraint that enforces explicit grounding of textual entities to corresponding visual regions. This eliminates the need for the model to infer correspondences from global representations, thereby facilitating efficient object-level alignment. 

To implement this, we develop a data synthesis pipeline to generate 773K high-quality samples with accurate object-entity correspondences. Given an image, our pipeline generates a detailed caption, extracts textual entities, and employs a grounding model to localize the corresponding visual objects. Empirically, MMCS demonstrates extraordinary data efficiency compared to standard image-level pretraining. With only 50K samples, our model achieves performance exceeding models pretrained on 600K standard image-caption pairs. Moreover, MMCS maintains consistent gains when scaling up both the dataset size and model capacity, yielding average improvements of 7.9\% on visual grounding and 2.1\% on perception-centric benchmarks. Further in-depth analysis confirms that these gains stem from higher-fidelity representation alignment and sharper attention distributions.

Our contributions are summarized as follows:
\begin{itemize}
    \item We introduce MultiModal Code-Switching (MMCS), a novel pretraining paradigm that interleaves visual objects into text, shifting alignment from implicit image-level associations to explicit object-level correspondences.
    \item We develop a scalable data synthesis pipeline that generates 773K samples with precise object-entity correspondences, bypassing the need for manual annotation.
    \item We conduct extensive experiments across various model scales and vision encoders, demonstrating the effectiveness of MMCS. We further provide mechanistic insights into how MMCS enhances the internal feature space topology of MLLMs.
\end{itemize}

\begin{figure*}[t]
    \centering
    \includegraphics[width=\textwidth]{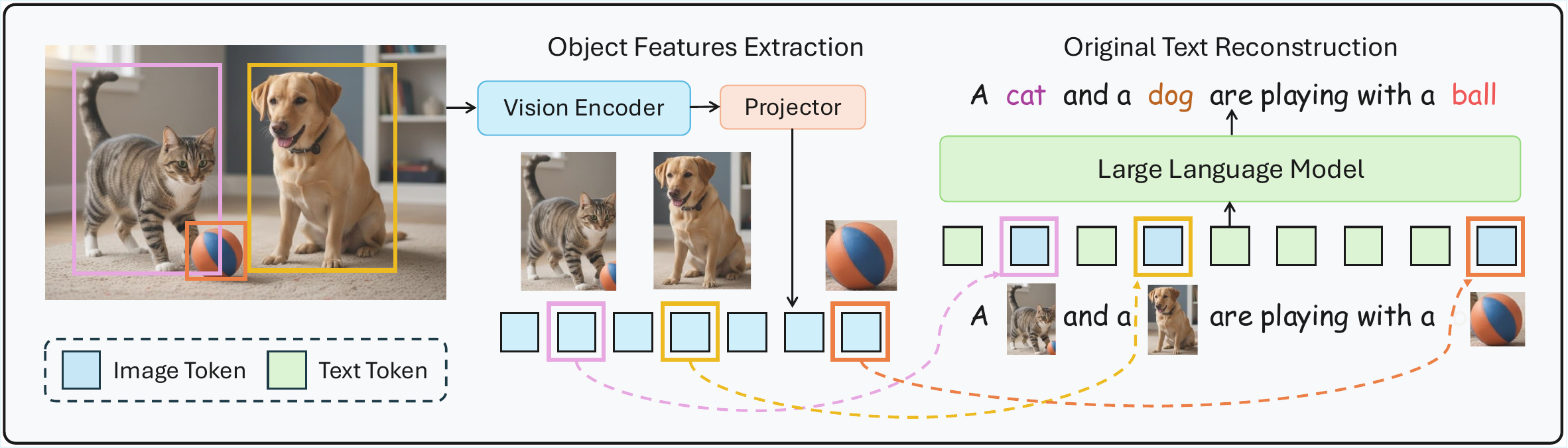}
    \caption{\textbf{Overview of MMCS pretraining paradigm.} We construct an interleaved image-text sequence by substituting textual entities with corresponding extracted visual objects. The generation of these entities and the subsequent context are conditioned on their visual counterparts, thereby facilitating object-level alignment.}
    \label{fig:mmcs}
\end{figure*}

\section{Related Works}
\label{sec:related_works}

\paragraph{Multimodal Large Language Models}

Current mainstream MLLMs adopt the ViT-MLP-LLM paradigm \citep{llava_onevision_15,internvl3,seed15vl}. Specifically, this architecture employs an MLP-based projector to align image features from a pretrained vision encoder with the input embedding space of an LLM backbone. These mapped visual tokens serve as soft prompts for conditional text generation. Training typically follows a two-stage strategy: the projector is first pretrained on image-caption pairs for modality alignment, followed by joint optimization of the projector and LLM on multimodal instruction data during SFT \citep{llava1.5,mplug_owl2}. The prevailing consensus in recent research is to utilize dense image captions during the modality alignment stage to provide detail-rich linguistic signals \citep{sharegpt4v,densefusion,molmo}. 

Despite these advancements, models trained on holistic, detailed captions still suffer from referential ambiguity: they must implicitly infer the correspondences between textual entities and their specific visual regions from a global image representation. Consequently, these models often struggle to disentangle individual visual concepts in complex scenes, relying instead on learned statistical co-occurrence patterns. This implicit alignment process leads to data inefficiency and suboptimal semantic grounding. In contrast, our approach provides explicit object-entity mappings, directly enforcing the grounding of textual descriptions in their corresponding visual regions.

\paragraph{Fine-Grained Multimodal Alignment}
The limitations of image-level alignment have motivated two lines of fine-grained methodologies. The first, patch-level alignment, pairs individual visual patches with textual labels through auxiliary objectives, such as contrastive loss over CLIP-assigned labels \citep{sea} or similarity maximization with vision-expert labels \citep{pat}. The second, grounding-capable MLLMs, establishes object-entity correspondences either via dedicated region-interaction modules \citep{glamm,lisa,ferret} or by encoding regions as textual coordinates or location tokens within the language interface \citep{shikra,kosmos2,groundinggpt}.

MMCS differs from both lines along distinct axes. Compared to patch-level methods, MMCS adopts complete visual objects as the alignment unit—patches rarely correspond to complete semantic units, yielding fragmented signals, whereas object-level supervision is cleaner and semantically coherent. Compared to grounding-capable MLLMs, which target task-level grounding capability through specialized modules or coordinate tokens, MMCS instead targets higher-fidelity representation-level alignment, introducing neither auxiliary modules nor coordinate supervision.

\section{Methods}
\label{sec:method}

\subsection{Insight: Code-Switching}

Our method draws inspiration from the linguistic phenomenon of code-switching, defined as the interleaving of two or more languages within a single utterance \citep{cs_2018}. For example, in the sentence ``The \textit{Klavier} is a versatile keyboard instrument,'' the German term \textit{Klavier} (piano) functions as a code-switched entity embedded within the English context.

Conceptually, code-switching implies that the switched term must be semantically compatible with its context. We leverage this principle by treating visual objects as distinct ``codes'' carrying specific semantic information. Just as a bilingual speaker selects appropriate words from either language, we substitute textual entities with their visual counterparts. This forces the model to resolve the visual representation to satisfy the semantic expectations of the sentence, thereby achieving explicit object-level grounding.

\subsection{Multimodal Code-Switching Pretraining}
\label{sec:mmcs}

Building on this insight, we introduce Multimodal Code-Switching (MMCS), a novel alignment paradigm that establishes correspondence between visual objects and textual entities. As illustrated in Figure \ref{fig:mmcs}, we construct interleaved code-switching image-text sequences by replacing the entity tokens with their corresponding visual objects. This substitution strategy creates a dependency chain: the generation of the subsequent textual context, as well as the reconstruction of the entity itself, is directly conditioned on the visual counterparts. Therefore, the model is compelled to ground entities to their corresponding visual regions, eliminating the need to infer such correspondences solely from global representations.

Formally, let $\mathbf{X}$ denote the original sequence of text tokens. Consider a contiguous textual entity segment $\mathbf{e}=[x^{i},\dots,x^{i+m}]$ starting at index $i$. Let $\mathbf{v}_{\text{object}}$ denote the visual representation of the corresponding object, defined as the set of image tokens whose spatial regions intersect with the object's bounding box. The interleaved image-text sequence $\mathbf{X}_\text{MMCS}$ is constructed by replacing the textual entity with the extracted object tokens:
\begin{equation}
    \mathbf{X}_{\text{MMCS}} = \text{Concat} (\mathbf{X}^{<i}, \mathbf{v}_{\text{object}}, \mathbf{X}^{>i+m}).
\label{eq:mmcs}
\end{equation}

We employ a language modeling objective where the loss is computed exclusively on text tokens, treating the visual tokens as conditioning context for next-token prediction:
\begin{equation}
    \mathcal{L}_\text{LM} = - \sum_{x_\text{MMCS}^t \in \text{Text}} \log p_\theta (x_\text{MMCS}^t \mid \mathbf{X}_{\text{MMCS}}^{<t}).
\end{equation}
To further enforce explicit supervision on object-entity correspondence, we additionally minimize the negative log-likelihood of the original textual entity segment given the preceding context:
\begin{equation}
    \mathcal{L}_{\text{entity}} = - \log p_\theta (\mathbf{e} \mid \mathbf{X}_{\text{MMCS}}^{<i}, \mathbf{v}_{\text{object}}),
\end{equation}
where $\mathbf{e}$ represents the substituted textual entity tokens in the original text sequence. The overall MMCS pretraining objective combines the language modeling loss and the entity reconstruction loss:
\begin{equation}
    \mathcal{L}_\text{MMCS} = \mathcal{L}_\text{LM} + \mathcal{L}_{\text{entity}}.
\label{eq:loss_mmcs}
\end{equation}

Since visual tokens replace the textual entity $e$ in $X_{\text{MMCS}}$, these two losses are non-overlapping and complementary. $\mathcal{L}_{\text{entity}}$ serves as the dedicated semantic anchor for entity reconstruction, while $\mathcal{L}_\text{LM}$ handles contextual integration.

\begin{figure}[t]
    \centering
    \includegraphics[width=0.45\textwidth]{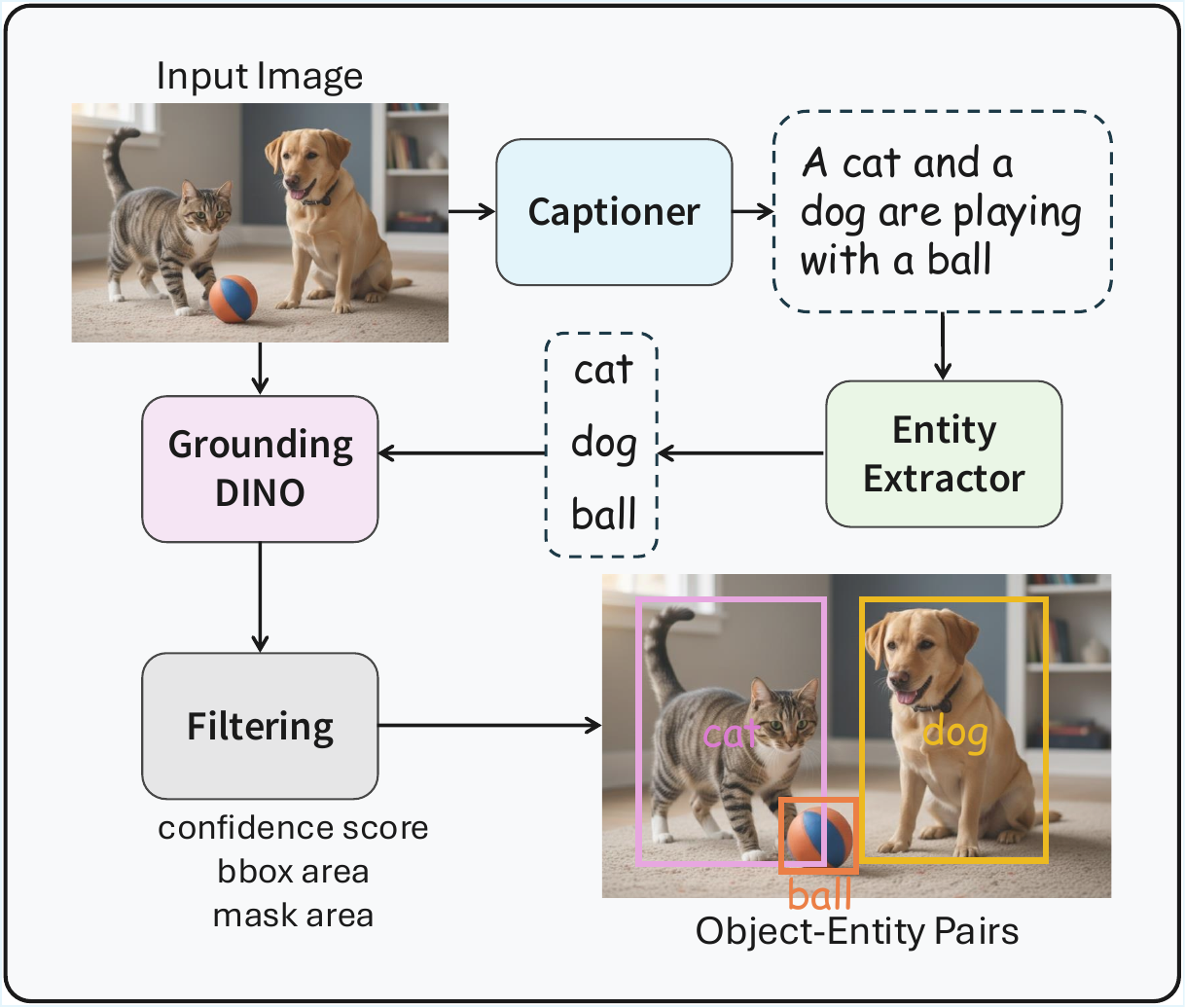}
    \caption{Data synthesis pipeline.}
    \label{fig:data}
\end{figure}

\begin{figure*}[t]
    \centering
    \includegraphics[width=1.0\textwidth]{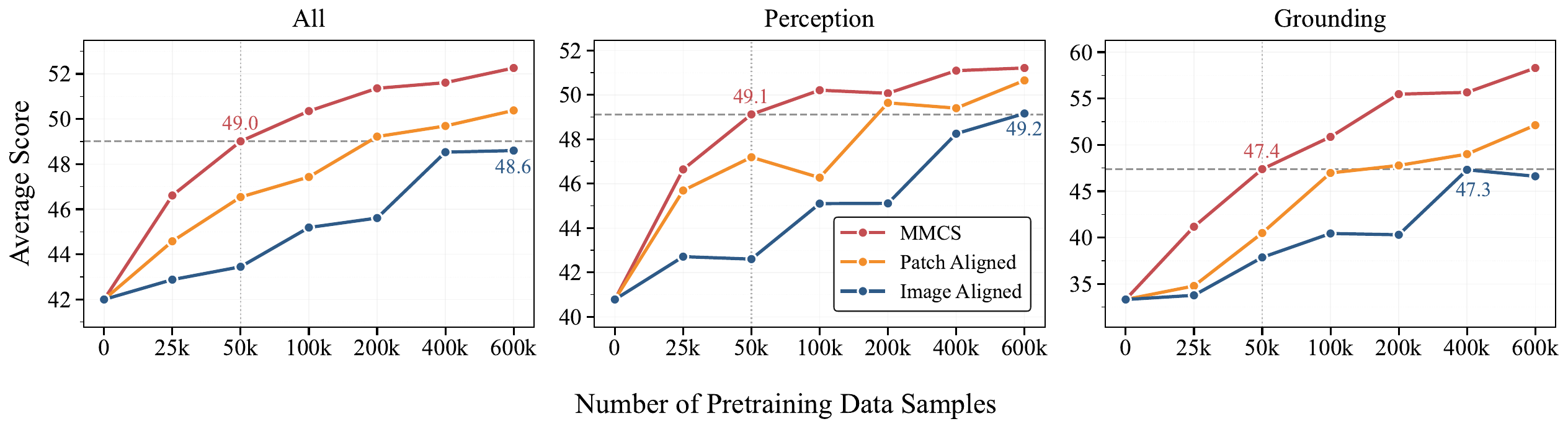}
    \caption{\textbf{Analysis of data efficiency.} We utilize Qwen2.5-3B as the LLM backbone and evaluate data efficiency by varying the pretraining dataset size from 0 to 600K, while keeping the SFT dataset fixed at 200K. The curves depict average scores across our benchmark suite, demonstrating that MMCS achieves superior data efficiency over both the Patch Aligned method and image-level pretraining (Image Aligned). Detailed compositions of the benchmarks used to calculate the average scores are provided in Appendix \ref{app:benchmarks}.}
    \label{fig:stage1_data_efficiency}
\end{figure*}

\subsection{Data Synthesis Pipeline}
\label{sec:data}
As shown in Figure \ref{fig:data}, we develop a data synthesis pipeline to generate interleaved code-switching image-text sequences with precise object-entity correspondence. For each input image, the pipeline proceeds through the following steps:

\paragraph{Detailed Image Captioning}
We begin by utilizing Qwen3-VL-32B-Instruct \citep{qwen3vl} to generate detailed image captions. To capture granular visual details, we prompt the caption model to exhaustively describe the attributes of all visible elements within the scene.

\paragraph{Textual Entity Extraction}
Given the generated detailed captions, we employ Qwen2.5-72B-Instruct \citep{qwen25} to extract textual entities. The output comprises a list of noun phrases, often accompanied by brief attribute descriptions (e.g., a white mug labeled ``O.CO''). These attributes are essential for disambiguating instances where multiple objects of the same category are present in a single image.

\paragraph{Visual Object Grounding}
We leverage Grounding DINO \citep{groundingdino} to anchor these textual entities to their corresponding visual regions. For each successfully localized entity, the output is a tuple containing the bounding box coordinates, the text label, and an associated confidence score.

\paragraph{Filtering}
To ensure high-quality, one-to-one object-entity correspondence, we implement a multi-stage filtering process. First, Grounding DINO predictions are filtered using a box threshold of 0.4 and a text threshold of 0.3. Subsequently, we calculate the area of each retained bounding box and prune those violating spatial constraints (e.g., smaller than a single visual patch or exceeding 50\% of the total image area). Finally, SAM-2.1 \citep{sam2} is employed to generate segmentation masks. Objects whose mask areas account for less than 20\% of their bounding-box areas are discarded to exclude heavily occluded instances.

We curate a diverse collection of images and apply the aforementioned synthesis pipeline. The generated pretraining dataset comprises 773K samples, each containing a source image, its detailed caption, and a set of visual objects paired with their bounding boxes and textual entity labels. A statistical summary and a detailed quality analysis of the synthesized dataset are presented in Appendix \ref{app:dataset}.

\begin{table*}[t]
    \centering
    \small
    \begin{tabular*}{0.95\textwidth}{@{\extracolsep{\fill}}llccccccccc}
    \toprule
    \multirow{2}{*}[-0.8ex]{LLM} & \multirow{2}{*}[-0.8ex]{Method} & \multicolumn{3}{c}{RefCOCO} & \multicolumn{3}{c}{RefCOCO+} & \multicolumn{2}{c}{RefCOCOg} & \multirow{2}{*}[-0.8ex]{AVG} \\
    \cmidrule(lr){3-5} \cmidrule(lr){6-8} \cmidrule(lr){9-10}
     & & testA & testB & val & testA & testB & val & test & val & \\
    \midrule
    
    \multirow{2}{*}{Qwen2.5 3B}
    & Caption & 65.00 & 54.64 & 61.68 & 63.88 & 47.49 & 56.62 & 61.27 & 62.39 & 59.12 \\
    & MMCS    & \textbf{80.66} & \textbf{68.75} & \textbf{75.18} & \textbf{74.35} & \textbf{56.23} & \textbf{65.49} & \textbf{70.32} & \textbf{71.31} & \textbf{70.29} \\
    \midrule
    
    \multirow{2}{*}{Qwen3 8B}   
    & Caption & \textbf{85.40} & 74.15 & 81.51 & 80.89 & 64.98 & 72.82 & 76.61 & 78.76 & 76.89 \\
    & MMCS    & 85.02 & \textbf{77.80} & \textbf{84.21} & \textbf{83.93} & \textbf{68.71} & \textbf{76.98} & \textbf{80.11} & \textbf{81.58} & \textbf{79.79} \\
    \midrule
    
    \multirow{2}{*}{Llama3 8B}  
    & Caption & 73.29 & 56.23 & 65.94 & 70.10 & 50.62 & 60.59 & 63.86 & 64.65 & 63.16 \\
    & MMCS    & \textbf{83.08} & \textbf{70.99} & \textbf{78.79} & \textbf{77.35} & \textbf{57.84} & \textbf{68.51} & \textbf{72.38} & \textbf{74.30} & \textbf{72.91} \\
    
    \bottomrule
    \end{tabular*}
    \caption{Performance on referring expression comprehension. ``Caption'' denotes image-level pretraining with image-caption pairs. We report Acc@0.5 across all splits.}
    \label{tab:refcoco}
\end{table*}

\begin{table*}[t]
\centering
\small
\begin{tabular*}{0.95\textwidth}{@{\extracolsep{\fill}}lll ccccccc}
\toprule

\multirow{2}{*}[-0.8ex]{LLM} & \multirow{2}{*}[-0.8ex]{ViT} & \multirow{2}{*}[-0.8ex]{Method} & \multicolumn{4}{c}{Perception} & \multicolumn{3}{c}{General} \\
\cmidrule(lr){4-7} \cmidrule(l){8-10}
& & & CVBench & OCR & VQA$^\text{T}$ & V* & MMB & MME & MMStar \\
\midrule

\multirow{2}{*}{Qwen2.5 3B} & \multirow{2}{*}{SigLIP2} & Caption & 61.26 & 44.70 & 58.29 & 48.17 & 67.53 & 60.36 & 42.77 \\
& & MMCS & \textbf{66.87} & \textbf{47.60} & \textbf{59.35} & \textbf{50.79} & \textbf{68.56} & \textbf{62.75} & \textbf{43.98} \\
\midrule

\multirow{2}{*}{Qwen2.5 3B} & \multirow{2}{*}{QwenViT} & Caption & 56.33 & 58.80 & 62.87 & 57.59 & 67.70 & 62.82 & 45.21 \\
& & MMCS & \textbf{59.78} & \textbf{62.40} & \textbf{64.97} & 57.59 & \textbf{68.47} & \textbf{65.68} & \textbf{45.31} \\
\midrule

\multirow{2}{*}{Qwen3 8B} & \multirow{2}{*}{SigLIP2} & Caption & 69.71 & 50.60 & 64.37 & 50.79 & 74.48 & 68.14 & 48.67 \\
& & MMCS & \textbf{72.82} & \textbf{53.30} & \textbf{65.23} & \textbf{53.40} & \textbf{76.63} & \textbf{68.54} & \textbf{51.49} \\
\midrule

\multirow{2}{*}{Llama3 8B} & \multirow{2}{*}{SigLIP2} & Caption & 65.66 & 47.00 & 62.59 & 51.31 & \textbf{70.79} & 63.43 & \textbf{44.96} \\
& & MMCS & \textbf{69.48} & \textbf{48.90} & \textbf{64.50} & \textbf{53.93} & 68.99 & \textbf{65.54} & 43.15 \\
\bottomrule

\end{tabular*}
\caption{Performance on perception-centric and general VQA benchmarks. ``OCR'', ``VQA$^\text{T}$'', ``V*'', and ``MMB'' denote OCRBench, TextVQA, V-Star, and MMBench, respectively. We report the normalized score for MME.}
\label{tab:perception_general}
\end{table*}

\section{Experiments}
\label{sec:experiments}

\subsection{Experiment Setup}
\label{sec:setup}

\paragraph{Model Architecture}
1) \textbf{Vision Encoder}: Unless otherwise specified, we use SigLIP2-SO-400M \citep{siglip2} equipped with a tile-wise dynamic resolution strategy \citep{llavanext}. 2) \textbf{LLM}: MMCS is integrated with Qwen2.5-3B-Instruct \citep{qwen25}, Qwen3-8B \citep{qwen3}, and Llama3-8B-Instruct \citep{llama3}. 3) \textbf{Projector}: We employ a two-layer MLP with a $2 \times 2$ pixel unshuffle operation \citep{internvl25} to reduce the number of image tokens.

\paragraph{Training Dataset}
For pretraining, we use the 773K dataset synthesized via the pipeline described in Section \ref{sec:data}. Crucially, both MMCS and the image-level baseline are trained on identical image–caption pairs, ensuring that any observed gains are attributable to our proposed approach rather than higher data quality. For SFT, we adopt the LLaVA-NeXT \citep{llavanext} dataset, which contains 779K high-quality instruction-following samples. We revert to using standard global image representations during SFT.

\paragraph{Evaluation}
Our evaluation covers a comprehensive suite of benchmarks categorized into three domains: 1) \textbf{Visual Grounding}: evaluating region understanding and grounding capabilities on referring expression comprehension (REC) tasks. 2) \textbf{Visual Perception}: including OCR-related tasks and real-world fine-grained perception. 3) \textbf{General VQA}: covering broad visual understanding and reasoning tasks.

For details on training setting and more evaluation results, please refer to Appendix \ref{app:implementation}.

\subsection{Analysis of Data Efficiency}
\label{sec:analysis_data_efficiency}

We first investigate the data efficiency of MMCS by varying the scale of the pretraining dataset. Specifically, we compare MMCS against standard image-level supervision and Patch Aligned Training \citep{pat} across dataset sizes ranging from 0 to 600K samples. To isolate the impact of pretraining, we use a constant subset of 200K SFT samples. As illustrated in Figure \ref{fig:stage1_data_efficiency}, MMCS exhibits superior data efficiency. With only 50K samples, MMCS achieves downstream performance that surpasses the image-level baseline trained on 600K samples. We attribute this advantage to the explicit object-entity supervision introduced in our method, which enables the model to efficiently align visual objects with their corresponding textual entities.

\subsection{Scaling Up Training Data and Model Sizes}
\label{sec:scale up}

To investigate the universality of MMCS, we scale up the training to utilize the full 773K pretraining and 779K instruction-following datasets. To verify robustness across different model sizes and architectures, we extend our evaluation to larger LLMs, including Qwen3-8B and Llama3-8B.

\paragraph{Visual Grounding} We evaluate visual grounding capabilities on referring expression comprehension tasks. As shown in Table \ref{tab:refcoco}, MMCS exhibits considerable improvements over the image captioning baseline across RefCOCO/+/g datasets, achieving an average gain of 7.9\%. Notably, since MMCS does not introduce bounding box coordinates of visual objects during pretraining, these improvements stem directly from enhanced object recognition and region understanding capabilities, highlighting the effectiveness of explicit object-level supervision during modality alignment.

\paragraph{Visual Perception} As shown in Table \ref{tab:perception_general}, MMCS consistently outperforms the standard image-captioning paradigm in perception tasks, maintaining robust gains as data volumes and model sizes increase. Our method demonstrates considerable performance gains in fine-grained visual perception tasks, achieving average improvements of 4.0\% on CVBench, 2.8\% on OCRBench and 2.0\% on V-Star. These results suggest that by grounding textual descriptions to specific visual regions, MMCS effectively facilitates precise visual perception.

\paragraph{General VQA} MMCS also improves performance on general VQA tasks, as shown in Table \ref{tab:perception_general}. We attribute this improvement to enhanced visual perception and region understanding, which provide more accurate visual evidence. The gain is more modest than those on grounding and perception because general VQA additionally depends on world knowledge, commonsense reasoning, and instruction-following priors that MMCS does not directly target. Nevertheless, our analysis in Appendix \ref{app:statistical_significance} shows that the improvement is statistically significant across multiple independent runs.

For more evaluation results, please refer to Appendix \ref{app:more_results}.

\subsection{Generalization Across Vision Encoders} Beyond tile-wise approaches, native dynamic resolution \citep{qwen2vl,qwen25vl} constitutes another prominent strategy for processing high-resolution images. This strategy encodes images into a variable number of visual tokens while preserving original aspect ratios \citep{navit}. We demonstrate our method's general effectiveness using QwenViT, the vision encoder utilized in Qwen2.5-VL \citep{qwen25vl}, which employs this native dynamic resolution mechanism. As evidenced in Table \ref{tab:perception_general}, MMCS maintains its superior performance in this setting. These results validate our method's compatibility across diverse vision encoders.

\begin{table}[t]
    \centering
    \small
    \setlength{\tabcolsep}{7.5pt}
    \begin{tabular}{lccc}
    \toprule
    Method & General & Perception & Grounding   \\
    \midrule
    Caption          & 51.19 & 55.77 & 59.12 \\
    SEA              & 50.78 & 55.69 & 60.96 \\
    Patch Aligned    & 51.78 & 56.43 & 66.55 \\
    Text BBox        & 51.64 & 56.62 & 64.44 \\
    MMCS (ours)      & \textbf{51.84} & \textbf{58.24} & \textbf{70.29} \\
    \midrule
    w/o $\mathcal{L}_\text{entity}$ & 51.45 & 54.23 & 67.55 \\
    w/o $\mathcal{L}_\text{LM}$     & 51.37 & 56.26 & 65.44 \\
    \bottomrule
    \end{tabular}
    \caption{Comparison with fine-grained alignment baselines and ablation on loss components. All methods and ablations are implemented under the same pretraining and SFT setup using Qwen2.5-3B-Instruct.}
    \label{tab:comparison_ablation}
\end{table}

\subsection{Comparison with Fine-grained Alignment Strategies}
We compare MMCS against three fine-grained alignment baselines without introducing additional modules under the same pretraining and SFT setup. SEA \citep{sea} and Patch Aligned Training \citep{pat} operate at the patch level, aligning individual visual patches with text tokens through auxiliary objectives. The third baseline, Text BBox, instantiates the textualized spatial representation paradigm of Section \ref{sec:related_works} \citep{shikra,kosmos2} at the pretraining stage. Specifically, each entity's bounding-box coordinates are appended as a text string after the entity in the caption, while retaining the standard image-caption training format.

As shown in Table \ref{tab:comparison_ablation}, MMCS consistently outperforms all three baselines across general, perception, and grounding tasks. The gains over SEA and Patch Aligned confirm that aligning semantically complete objects yields cleaner supervision. Text BBox shares MMCS's principle of explicit object-entity correspondence, but it conveys correspondence symbolically through coordinate strings, requiring the model to resolve tokens into spatial regions before reaching visual content. MMCS instead binds visual objects to textual entities at the representation level, and this binding proves more effective for modality alignment than symbolic coordinate cues.

See Appendix \ref{app:benchmarks} for the benchmark compositions of these average scores.

\subsection{Ablation Study}
We perform an ablation study on the components of Eq. \ref{eq:loss_mmcs}. The results in Table \ref{tab:comparison_ablation} demonstrate that both objectives are crucial for optimal performance.

\paragraph{$\mathcal{L}_{\text{entity}}$ enforces semantic precision.} 
Removing the entity reconstruction loss $\mathcal{L}_{\text{entity}}$ leads to the most pronounced drop on perception benchmarks (-4.0\%). This indicates that $\mathcal{L}_{\text{entity}}$ acts as a \textit{semantic anchor}. By requiring the model to translate visual object embeddings back into their corresponding textual entities, it compels the projector to encode discriminative, identity-bearing visual features, thereby enhancing visual perception.

\paragraph{$\mathcal{L}_{\text{LM}}$ facilitates contextual integration.}
Removing the language modeling loss $\mathcal{L}_{\text{LM}}$, in contrast, causes the largest drop on grounding benchmarks (-4.9\%). We attribute this to its role in \textit{contextual integration}. It trains the model to predict the textual context surrounding each visual object, encouraging it to reason about how the object relates to its descriptive attributes and to other entities in the scene, which is critical for both visual perception and referring expression comprehension.

\section{Discussion}
\label{sec:discussion}

\begin{figure*}[t]
  \includegraphics[width=\linewidth]{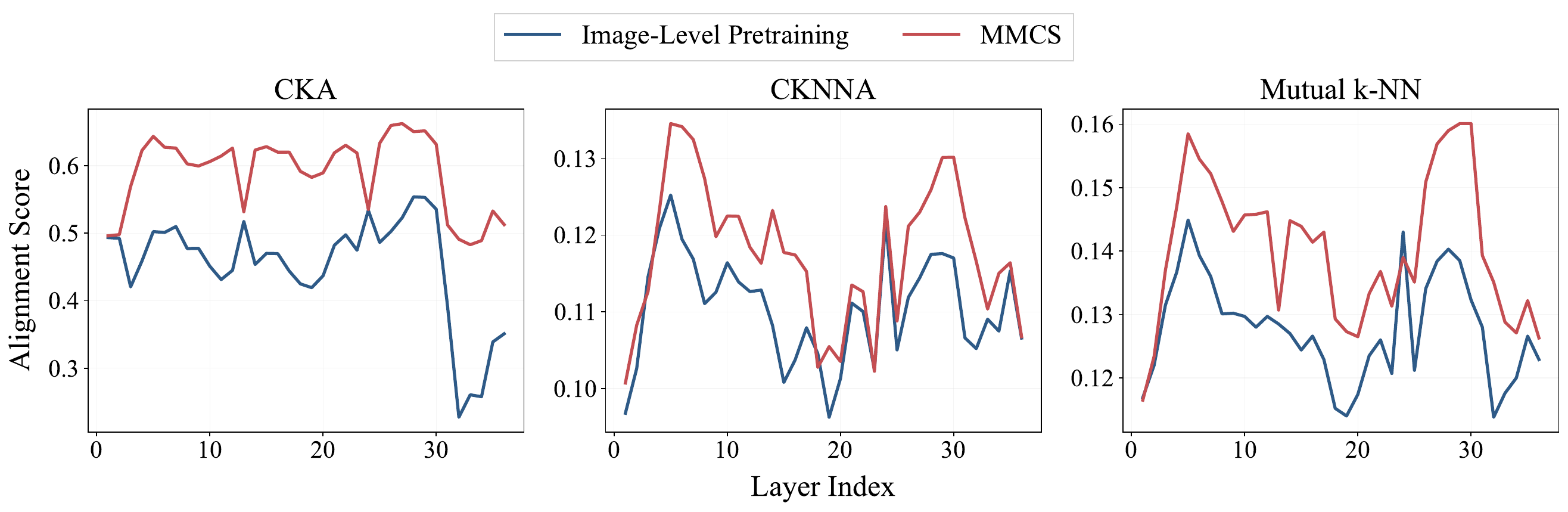} 
  \caption{\textbf{Layer-wise representation alignment.} To evaluate cross-modal integration, we extract hidden states from each LLM decoder layer for input image-text pairs and partition them by modality. We then quantify the alignment using three distinct metrics: CKA, CKNNA and Mutual k-NN. MMCS achieves better representation alignment than standard image-level pretraining across most layers.}
  \label{fig:layer_wise_alignment}
\end{figure*}

\begin{figure*}
    \centering
    \includegraphics[width=1.0\textwidth]{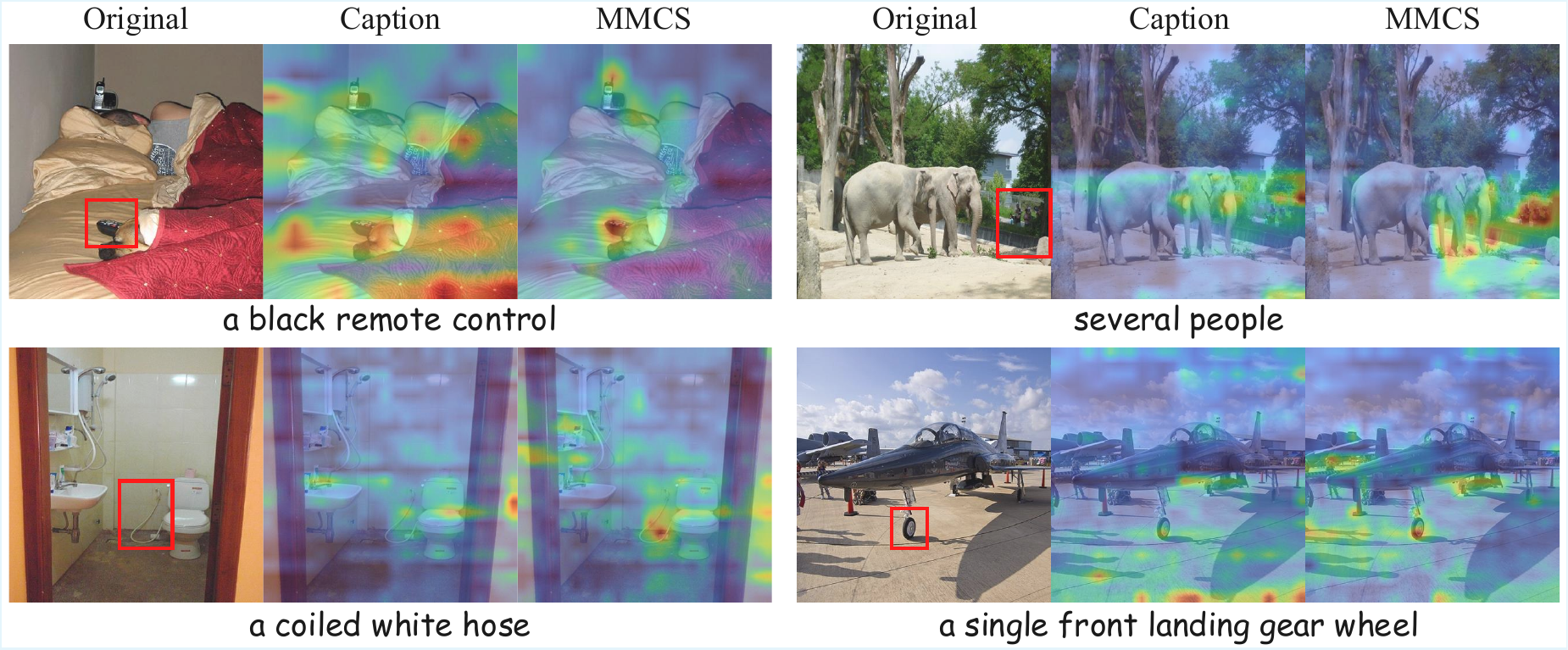}
    \caption{\textbf{Visualization of attention maps.} Each example displays a triplet containing: (Left) the original image with a red bounding box highlighting the target object; (Middle) the attention map from an MLLM pretrained with standard captioning; and (Right) the attention map from our MMCS method. Warmer colors indicate higher attention weights (min-max normalized). The specific text entity driving the attention is noted below each example.}
    \label{fig:attn_vis}
\end{figure*}

\subsection{Representation Alignment Measurement}
\label{sec:representation_alignment}
To investigate whether MMCS achieves superior multimodal alignment at the feature level compared to standard image-level pretraining, we directly quantify vision-language representational alignment after pretraining. We employ three metrics: CKA \citep{cka} to evaluate \textit{global} geometric correspondence, and Mutual k-NN and CKNNA \citep{platonic} to assess \textit{local} neighborhood consistency. Detailed definitions of these metrics are provided in Appendix \ref{app:alignment_metrics}.

Specifically, we input 1,000 image-caption pairs sampled from the COCO2014 validation set into the MLLM and extract hidden states from all layers. These states are partitioned into visual and textual components to compute layer-wise alignment metrics. As shown in Figure \ref{fig:layer_wise_alignment}, MMCS achieves superior representational alignment compared to image-level supervision. These findings align with the hypothesis that representational alignment correlates with model capability \citep{platonic}, offering a rationale for the performance improvements observed in downstream tasks. 

\subsection{Attention Map Analysis}

Figure \ref{fig:attn_vis} presents a qualitative comparison of attention maps. We analyze the cross-modal attention distribution from textual entity tokens to visual features by feeding image-caption pairs into the MLLM. The model pretrained with MMCS accurately attends to visual regions corresponding to specific textual entity descriptions. In contrast, the model trained with image-caption pairs often produces diffuse or misaligned attention patterns. Since the LLM backbone is frozen during pretraining, these results show that training the projector with explicit object–entity correspondence yields more semantically precise and interpretable features. These features align better with the LLM's pre-existing semantic space.

Further analyses on model capabilities, robustness and scalability of our approach, and additional qualitative results are provided in the Appendix \ref{app:extended_analysis}, \ref{app:robustness_scale} and \ref{app:more_attn_results}, respectively.

\section{Conclusion}
In this paper, we propose MMCS, a modality alignment pretraining paradigm for explicit object-entity correspondence. Extensive experiments show that MMCS considerably improves both data efficiency and downstream performance. We further analyze how MMCS facilitates multimodal representation alignment through internal feature space topology. Our findings highlight the importance of object-level alignment in developing data-efficient MLLMs with advanced performance.

\section*{Limitations}

While MMCS demonstrates considerable improvements in data efficiency and downstream performance, our current implementation primarily focuses on visual objects within natural images. The core methodology of establishing explicit correspondence between local visual features and textual entities is generalizable to broader scenarios. Extending MMCS to domains such as chart understanding or scene text recognition will be further explored in our future work. Furthermore, exploring bootstrapping mechanisms for iterative self-improvement and extending the MMCS paradigm to encompass complex relations and actions represent highly promising directions that we intend to investigate.

\paragraph{Dataset licenses and privacy.}
We use images from established research datasets but do not redistribute them. Our release contains only newly generated captions and object-entity annotations. All source materials remain subject to their original terms. COCO and GQA annotations are licensed under CC BY 4.0, while their images retain the applicable source licenses. Flickr30K images retain their original copyrights and are provided for non-commercial research and educational use. Objects365 annotations are licensed under CC BY 4.0, while its image rights are source-specific. Open Images lists its images under CC BY 2.0 and its annotations under CC BY 4.0. SA-1B is governed by the SA-1B Dataset Research License. Our release grants no additional rights to the underlying images, which users must obtain from the official providers. We introduce no newly scraped images or identity labels. However, the source images may contain faces, license plates, or location cues, and our generated annotations may reflect such information. The released annotations may therefore inherit privacy risks from the source datasets.

\paragraph{Bias and potential misuse.}
The synthesized data may inherit biases from both its source datasets and the automatic captioning and grounding models. Its object-category coverage may be uneven, particularly for rare, culturally specific, or otherwise underrepresented objects, potentially resulting in unequal grounding accuracy across categories and contexts. Moreover, stronger object-localization capabilities could be combined with identification systems for surveillance or tracking. We discourage such uses and recommend source-specific data filtering, domain-specific risk assessments, and appropriate safeguards before deployment.

% \section*{Acknowledgments}

% This document has been adapted
% by Steven Bethard, Ryan Cotterell and Rui Yan
% from the instructions for earlier ACL and NAACL proceedings, including those for
% ACL 2019 by Douwe Kiela and Ivan Vuli\'{c},
% NAACL 2019 by Stephanie Lukin and Alla Roskovskaya,
% ACL 2018 by Shay Cohen, Kevin Gimpel, and Wei Lu,
% NAACL 2018 by Margaret Mitchell and Stephanie Lukin,
% Bib\TeX{} suggestions for (NA)ACL 2017/2018 from Jason Eisner,
% ACL 2017 by Dan Gildea and Min-Yen Kan,
% NAACL 2017 by Margaret Mitchell,
% ACL 2012 by Maggie Li and Michael White,
% ACL 2010 by Jing-Shin Chang and Philipp Koehn,
% ACL 2008 by Johanna D. Moore, Simone Teufel, James Allan, and Sadaoki Furui,
% ACL 2005 by Hwee Tou Ng and Kemal Oflazer,
% ACL 2002 by Eugene Charniak and Dekang Lin,
% and earlier ACL and EACL formats written by several people, including
% John Chen, Henry S. Thompson and Donald Walker.
% Additional elements were taken from the formatting instructions of the \emph{International Joint Conference on Artificial Intelligence} and the \emph{Conference on Computer Vision and Pattern Recognition}.

\bibliography{custom}
\clearpage

\appendix

\section{Dataset Details}
\label{app:dataset}

\subsection{MMCS Pretraining Dataset}

\begin{table}[h]
    \centering
    \small
    \setlength{\tabcolsep}{7pt}
    \begin{tabular}{cccc}
        \toprule
        \# Samples & \# Objects & Avg. Chars & Avg. Objects \\
        \midrule
        773,779 & 5,145,630 & 961.65 & 6.65 \\
        \bottomrule
    \end{tabular}
    \caption{\textbf{Statistics of the synthesized pretraining dataset.} ``Avg. Chars'' denotes the average character count per detailed caption, while ``Avg. Objects'' represents the average number of visual objects identified per sample after filtering.}
    \label{tab:data_statistics}
\end{table}

\begin{table*}[t]
    \centering
    \small
    \setlength{\tabcolsep}{11pt}
    \begin{tabular}{l|ccccccc}
    \toprule
    Judge & COCO & Flickr30k & GQA & Objects365 & OpenImages & SA-1B & AVG \\
    \midrule
    VLM & 0.935 & 0.920 & 0.965 & 0.980 & 0.970 & 0.925 & 0.949 \\
    Human & 0.950 & 0.900 & 0.900 & 1.000 & 1.000 & 0.900 & 0.942 \\
    \bottomrule
    \end{tabular}
    \caption{Quality evaluation of the synthesized MMCS pretraining dataset. Both the VLM judge and human annotators assign binary scores indicating whether a textual entity is accurately grounded by its corresponding visual bounding box. The reported values represent the average grounding accuracy.}
    \label{tab:data_quality}
\end{table*}

\begin{figure*}[t]
    \centering
    \includegraphics[width=\textwidth]{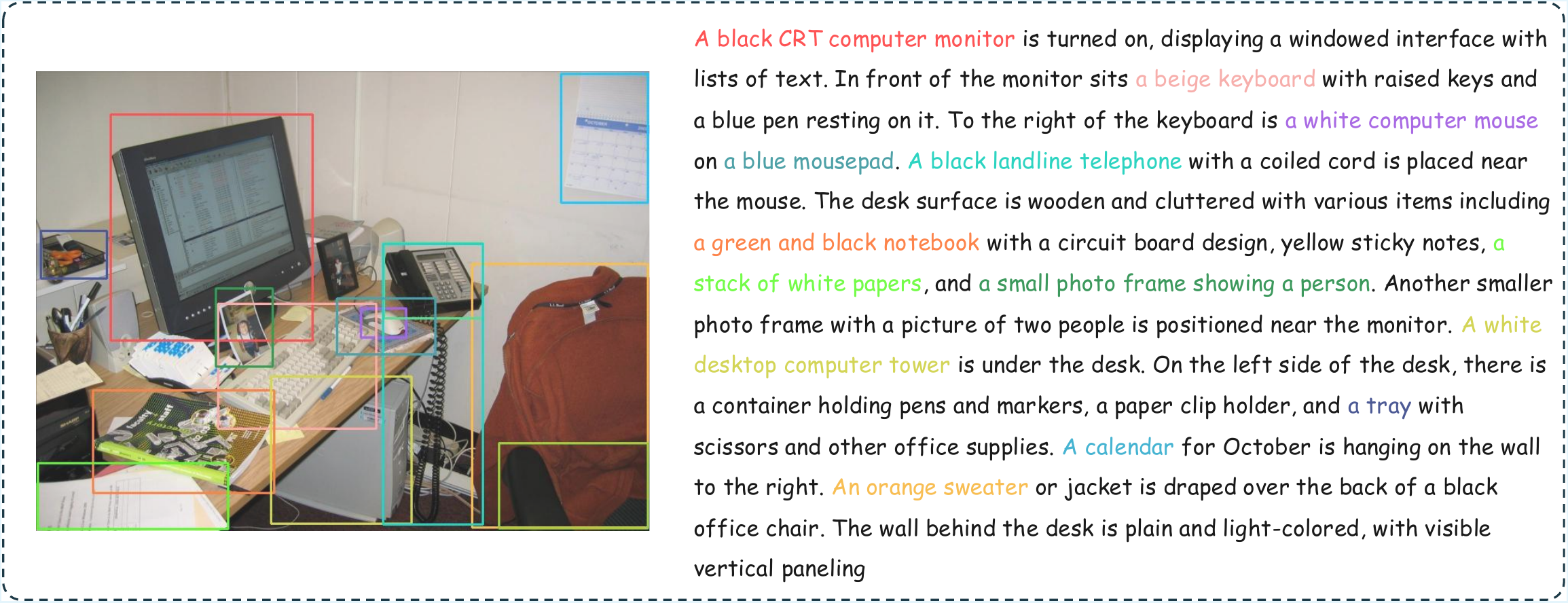}
    \includegraphics[width=\textwidth]{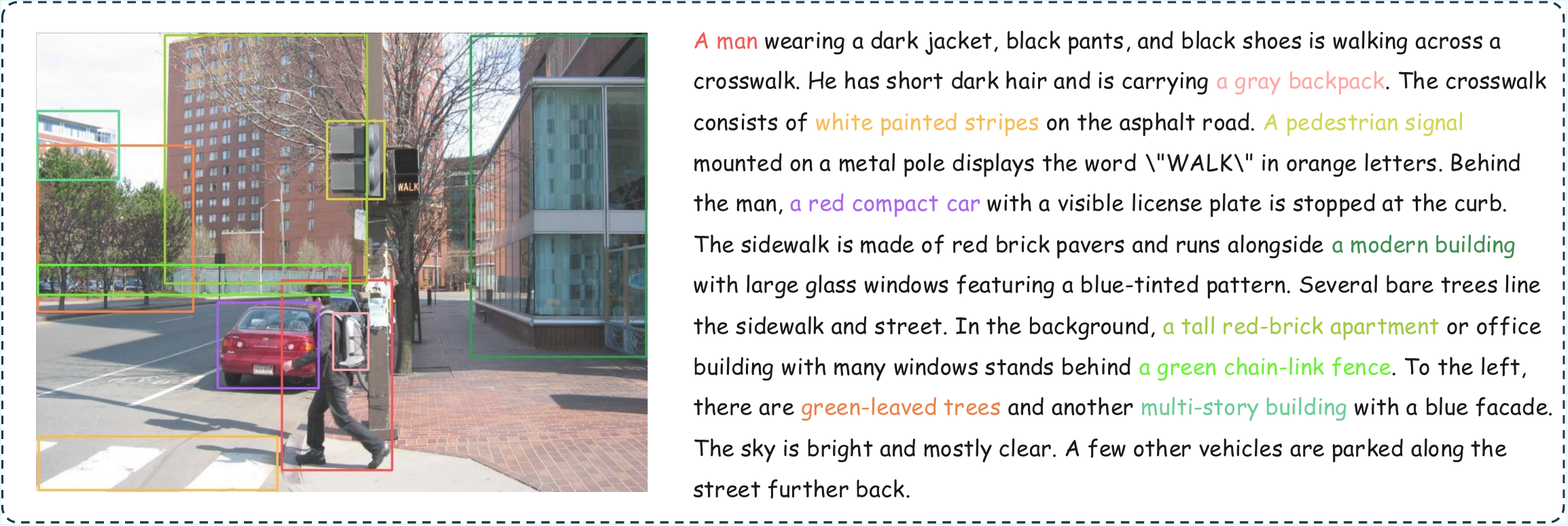}
    \caption{Illustrative examples from the pretraining dataset. \textbf{Left}: The original image with annotated bounding boxes of visual objects. \textbf{Right}: The dense caption of the image, in which successfully localized textual entities are highlighted. The textual entities are color-coded to match the corresponding grounded visual objects.}
    \label{fig:data_examples}
\end{figure*}

The image sources utilized in Section \ref{sec:data} include MS COCO \citep{coco}, Flickr30k \citep{flickr30k}, GQA \citep{gqa}, Objects365 \citep{obj365}, OpenImages \citep{openimages}, and SA-1B \citep{sam}. We select 773K images from these sources and apply our data synthesis pipeline to construct the pretraining dataset. Table \ref{tab:data_statistics} presents a statistical summary of the dataset, and Figure \ref{fig:data_examples} provides two illustrative examples.

\begin{tcolorbox}[colback=gray!5!white, colframe=gray!75!black, title=Prompts for VLM Judge]
\label{prompts_vlm_judge}
You are judging whether a bounding box correctly localizes the target described by a phrase. The image contains a highlighted bbox. Decide if it covers the intended object. \\

Target phrase: \{phrase\} \\
Bbox (xyxy): [\{x1, y1, x2, y2\}] \\
Image size (w×h): [\{width, height\}] \\

Criteria:

- YES: the highlighted region matches the phrase, and the target object is mostly inside the box.

- NO: the box highlights the wrong object or attribute, or the target is clearly outside the box. \\

Answer exactly: YES or NO

\end{tcolorbox}

\paragraph{Data Quality}
To validate the quality of our synthesized pretraining dataset, we use Gemini-3-Pro \citep{gemini3} as an automated Vision-Language Model (VLM) judge and conduct an independent human evaluation to verify the reliability of its assessments. For each source dataset, we sample 200 objects to compute the VLM score. Specifically, the judge is prompted to assign a binary score (1 for correct, 0 for incorrect) indicating whether a textual entity is accurately localized by the provided bounding box. To assess the reliability of the VLM evaluations, human annotators independently verify a subset of 20 randomly selected objects per dataset. The evaluation results, representing the average grounding accuracy, are summarized in Table \ref{tab:data_quality}.

\begin{tcolorbox}[colback=gray!5!white, colframe=gray!75!black, title=Instruction for Human Annotators]
\label{instructions_human_judge}
Task:

Each sample shows an image with a colored bounding box and a target phrase. Judge whether the box correctly highlights the object described by the phrase. \\

Label: YES or NO \\

YES: acceptable bbox \\
  - The box covers the correct object matching the phrase. \\
  - The object is mostly inside the box (minor over-/under-coverage is fine). \\
  - Slight attribute ambiguity is tolerable if the object identity is correct.
    e.g. phrase="red car", the car is dark red — still YES. \\

NO: wrong bbox \\
  - Wrong object: the box highlights a different object than what the phrase describes. \\
  - Wrong attribute: any key attribute clearly conflicts. \\
  - Missed target: the described object is mostly outside the box.
\end{tcolorbox}

\subsection{SFT Dataset}

For SFT, we utilize the LLaVA-NeXT dataset \citep{llavanext}, which comprises 779K instruction-following samples covering general VQA, OCR-related tasks and document/chart understanding. Specifically, this dataset incorporates data from AI2D \citep{ai2d}, ChartQA \citep{chartqa}, DocVQA \citep{docvqa}, DVQA \citep{dvqa}, GQA, LAION-GPT4V \citep{laion}, OCR-VQA \citep{ocr-vqa}, ShareGPT4V \citep{sharegpt4v}, SynthDoG-EN \citep{syndog} and Visual Genome \citep{vg}.

\subsection{Benchmarks}
\label{app:benchmarks}

We conduct a comprehensive evaluation across a suite of benchmarks organized into three distinct domains:

\begin{itemize}
    \item \textbf{Visual Grounding}: We evaluate referring expression comprehension capability using RefCOCO, RefCOCO+ \citep{refcoco}, and RefCOCOg \citep{refcocog}.
    \item \textbf{Perception-centric tasks}: We employ AI2D and ChartQA for diagram and chart understanding; OCRBench \citep{ocrbench} and TextVQA \citep{textvqa} to assess OCR capabilities; and CVBench \citep{cvbench}, RealWorldQA \citep{realworldqa}, and V-Star \citep{vstar} for real-world visual perception.
    \item \textbf{General VQA}: We encompass MMBench \citep{mmbench}, MME \citep{mme}, MMStar \citep{mmstar}, and MMVet \citep{mmvet} for general visual understanding; MMMU \citep{mmmu} for multi-disciplinary reasoning requiring college-level knowledge; and GQA \citep{gqa} for real-world compositional visual reasoning.
\end{itemize}

To ensure reproducibility, all evaluations are conducted using the lmms-eval framework \citep{lmms_eval}. The full evaluation results are in Table \ref{tab:full_grounding}, Table \ref{tab:full_perception}, and Table \ref{tab:full_general}.

\section{Implementation Details}
\label{app:implementation}

\begin{table}[t]
    \centering
    \small
    \setlength{\tabcolsep}{10pt}
    \begin{tabular}{l|cc}
        \toprule
         Hyperparameter & Pretrain & Finetune \\
         \midrule
         Trainable & MLP & MLP+LLM \\
         Global batch size     & 128      & 64       \\
         Learning rate  & 1e-3     & 2e-4     \\
         Epochs         & \multicolumn{2}{c}{1} \\
         Max image tiles & \multicolumn{2}{c}{4~+~1} \\
         Optimizer & \multicolumn{2}{c}{AdamW} \\
         LR schedule & \multicolumn{2}{c}{Cosine decay} \\
         Warm up ratio & \multicolumn{2}{c}{0.03} \\
         Weight decay & \multicolumn{2}{c}{0} \\
         LoRA r & - & 32 \\
         LoRA alpha & - & 64 \\
         LoRA dropout & - & 0.05 \\
         \bottomrule
    \end{tabular}
    \caption{\textbf{Hyperparameters for model training.} ``4+1'' indicates that the high-resolution image is divided into at most 4 localized tiles with an additional thumbnail tile that provides a global view.}
    \label{tab:hyper}
\end{table}

Table \ref{tab:hyper} details the hyperparameters employed for both modality alignment pretraining and SFT stages. High-resolution images are divided into smaller image tiles of the resolution that the ViT is originally trained for, and encoded independently. All experiments are conducted on 8 NVIDIA A6000 GPUs. During the SFT stage, we apply LoRA \citep{lora} to the LLM backbone. For 8B-scale models, pretraining completes within 10 hours, while SFT completes within 20 hours.

\section{Extended Analysis}
\label{app:extended_analysis}

\subsection{Learning of Actions and Relationships}

While MMCS explicitly grounds noun phrases, we hypothesize that its learning signal is not limited to nouns. Since the visual object tokens are inserted into the surrounding sentence context, the language modeling objective also supervises non-entity tokens such as verbs and prepositions. Therefore, actions and relations can be learned through their contextual dependence on the grounded visual objects. To verify this, we evaluate how well the model predicts different types of text tokens conditioned on the entire image. Given an image $I$ and a caption token sequence $X=\{x_1,\dots,x_N\}$, we compute the token-level prediction probability $p(x_i | I, x_{<i})$, and report the average negative log-likelihood for each part-of-speech category $C$:
\begin{equation}
    L_C = -\frac{1}{|S_C(X)|} \sum_{i \in S_C(X)} \log p(x_i | I, x_{<i}),
\end{equation}
where $S_C(X)$ is the set of token positions in caption $X$ that belong to category $C$. Lower $L_C$ indicates that the model predicts the corresponding tokens with higher probability, suggesting better understanding of that semantic category. Specifically, we sample 1,000 image-caption pairs from the ShareGPT-4V \citep{sharegpt4v} dataset and use spaCy (\texttt{en\_core\_web\_sm}) to group tokens into verbs, prepositions, and nouns, corresponding to actions, relationships, and objects, respectively. All evaluated models are pretraining-only checkpoints without undergoing SFT.

As reported in Table \ref{tab:pos_nll}, although MMCS only substitutes noun-phrase entities, it also yields notable NLL reductions on verbs and prepositions. Since these non-entity tokens are not directly replaced, they receive visual supervision only through the LM loss over the surrounding context, indicating that explicit object-level grounding implicitly propagates to actions and relations. By contrast, Patch Aligned improves nouns but barely affects verbs or prepositions (-0.01). Patch-level supervision aligns visual fragments with disconnected labels, offering little signal on \textit{how} objects act or relate. MMCS instead grounds complete objects within their natural sentence context, providing alignment signals on both concrete entities and the abstract concepts that bind them.

\begin{table}[t]
    \centering
    \small
    \begin{tabular}{l c c c c}
    \toprule
    Method & Verb & Prep & Noun & All \\
    \midrule
    Caption       & 3.4773 & 1.1833 & 2.2005 & 1.8097 \\
    Patch Aligned & 3.4660 & 1.1711 & 1.9974 & 1.7309 \\
    MMCS          & 3.2645 & 1.0845 & 1.9222 & 1.6362 \\
    \bottomrule
    \end{tabular}
    \caption{Token-level negative log-likelihood ($L_C$) by part-of-speech category. Verbs, prepositions, and nouns correspond to actions, relationships, and objects, respectively.}
    \label{tab:pos_nll}
\end{table}

\begin{table*}[t]
    \small
    \setlength{\tabcolsep}{11pt}
    \centering
    \begin{tabular}{lcccccc}
    \toprule
    Metric & COCO & Flickr30k & GQA & Objects365 & OpenImages & SA-1B \\
    \midrule
    \# Has Text & 91 & 50 & 107 & 82 & 119 & 154 \\
    Ratio (\%) & 18.2 & 10.0 & 21.4 & 16.4 & 23.8 & 30.8 \\
    \bottomrule
    \end{tabular}
    \caption{Proportion of visual objects containing legible text across different source datasets. Statistics are based on a random sample of 500 objects per dataset, evaluated by a VLM judge.}
    \label{tab:ocr_analysis}
\end{table*}

\subsection{Emergent OCR Capability}

As shown in Table \ref{tab:perception_general}, MMCS improves performance on OCR-related benchmarks (e.g., OCRBench, TextVQA) despite the absence of dedicated OCR pretraining data. We attribute this emergent capability to two primary factors:

\paragraph{Implicit OCR Supervision from Natural Objects}
Natural images frequently contain objects with embedded text (e.g., street signs, product labels). While standard image-level pretraining heavily compresses these small regions into global features, MMCS explicitly crops and aligns them with textual descriptions, acting as an implicit OCR training objective. To quantify this, we utilized Gemini-3-Pro to evaluate 500 random object crops per source dataset. As shown in Table \ref{tab:ocr_analysis}, a notable proportion of natural objects (10.0\% to 30.8\%) contain legible text, providing substantial implicit supervision.

\paragraph{Generalization via Enhanced Modality Alignment}

By explicitly resolving the referential ambiguity inherent in global image-text pairs, MMCS establishes a more precise modality alignment. This fundamental improvement enables the model to accurately localize and parse fine-grained visual details, a capability that naturally generalizes to text recognition tasks.

\subsection{Token Efficiency and Training Cost}
\label{sec:cost_comparison}

\begin{figure}[t]
    \centering
    \includegraphics[width=\linewidth]{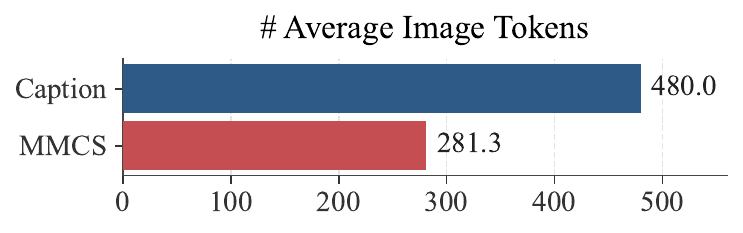}
    \caption{Comparison of the average number of image tokens between MMCS and standard image-level pretraining.}
    \label{fig:cost_comparison}
\end{figure}

In standard image-level pretraining, the entire image is encoded into a fixed set of image tokens. In contrast, MMCS operates explicitly on localized visual objects. As defined in Eq. \ref{eq:mmcs}, only the image patches that spatially intersect with the target bounding boxes are retained and fed into the LLM backbone, inherently reducing the input sequence length. To quantify this, we randomly selected 3,000 instances from each source dataset and compared the average number of image tokens per sample under identical dynamic resolution configurations. As shown in Figure \ref{fig:cost_comparison}, MMCS reduces the token count by approximately 41.4\% compared to standard image-level pretraining.

\section{Robustness and Scalability Analysis}
\label{app:robustness_scale}

\subsection{Robustness to Caption Quality}

A natural concern is whether the gains of MMCS stem from the code-switching formulation itself or from the quality of our teacher-generated captions. To disentangle these factors, we randomly sample 500K image-caption pairs from ShareGPT-4V \citep{sharegpt4v}, apply the same MMCS pipeline (entity extraction, grounding, and substitution) directly on top of these existing captions, and compare against a standard image-caption baseline trained on the same data. As shown in Table \ref{tab:robustness_caption_noise}, both methods exhibit lower absolute scores than our default setup, but MMCS still considerably outperforms the caption baseline. This confirms that the improvement arises from the MMCS formulation itself rather than from the specific teacher pipeline.

\subsection{Robustness to Noisy Object-Entity Correspondence}

Since our object-entity correspondences are produced by an automated grounding pipeline, they inevitably contain mismatches. To assess the robustness of MMCS to such errors, we randomly replace a proportion (10\% and 30\%) of the correct bounding boxes with irrelevant regions sampled from the same image, constrained to be of comparable size (80\%--125\% of the original) and to have minimal overlap with other valid objects (IoU $<0.1$). As shown in Table \ref{tab:robustness_caption_noise}, under 10\% perturbation MMCS remains highly robust, still surpassing the caption baseline across all task categories. Even under the more aggressive 30\% corruption, the model does not collapse and continues to deliver clear gains on grounding, while general and perception scores stay close to the baseline. These results show that MMCS does not require pixel-perfect object localization to be effective, making it amenable to scaling with imperfect automatic annotations.

\begin{table}[t]
  \centering
  \small
  \setlength{\tabcolsep}{3.2pt}
  \begin{tabular}{lccc}
    \toprule
    Method & General & Perception & Grounding \\
    \midrule
    Caption        & 51.19 & 55.77 & 59.12 \\
    MMCS           & 51.84 & 58.24 & 70.29 \\
    \midrule
    \rowcolor{gray!15}
    \multicolumn{4}{l}{\textit{Varying caption source}} \\
    \addlinespace[3pt]
    Caption (ShareGPT-4V)     & 49.35 & 53.64 & 58.70 \\
    MMCS (ShareGPT-4V)        & 52.16 & 56.96 & 65.55 \\
    \midrule
    \rowcolor{gray!15}
    \multicolumn{4}{l}{\textit{Perturbing object-entity correspondence}} \\
    \addlinespace[3pt]
    MMCS (10\% pert.)         & 51.43 & 56.93 & 66.44 \\
    MMCS (30\% pert.)         & 50.75 & 54.83 & 62.86 \\
    \bottomrule
  \end{tabular}
  \caption{Robustness to varying caption source and noisy object-entity correspondences.}
  \label{tab:robustness_caption_noise}
\end{table}

\begin{table}[t]
\centering
\small
\setlength{\tabcolsep}{4pt}
\begin{tabular}{llcccc}
\toprule
Max Res. & Method & General & Perception & Grounding \\
\midrule
\multirow{2}{*}{$384^2 \times 1$}
& Caption & 46.62 & 38.32 & 41.78 \\
& MMCS     & 49.45 & 42.75 & 44.76 \\
\midrule
\multirow{2}{*}{$384^2 \times 4$}
& Caption & 48.95 & 49.16 & 46.61 \\
& MMCS     & 50.47 & 51.21 & 58.29 \\
\midrule
\multirow{2}{*}{$384^2 \times 9$}
& Caption & 48.70 & 50.51 & 46.05 \\
& MMCS     & 50.86 & 53.08 & 58.96 \\
\midrule
\multirow{2}{*}{$384^2 \times 12$}
& Caption & 49.00 & 48.36 & 50.44 \\
& MMCS     & 50.73 & 54.39 & 58.07 \\
\bottomrule
\end{tabular}
\caption{Performance comparison across \textbf{different maximum image resolutions}. The input resolution is dynamically scaled by varying the maximum number of $384 \times 384$ tiles. MMCS consistently outperforms the standard image-level pretraining baseline.}
\label{tab:resolution}
\end{table}

\subsection{Robustness to Image Resolution}
We evaluate the robustness of MMCS across varying input image resolutions. By adjusting the maximum number of tiles (ranging from 1 to 12), we compare our method against the standard caption baseline with a fixed set of 600K pretraining samples and 200K SFT samples. As shown in Table \ref{tab:resolution}, higher dynamic resolutions generally yield better performance in perception and grounding tasks. Crucially, MMCS consistently outperforms the baseline across all resolution configurations, underscoring the efficacy of our method regardless of the visual input granularity.

\begin{table}[t]
\centering
\setlength{\tabcolsep}{4pt}
\small
\begin{tabular}{llccc}
\toprule
\# Samples & Method & General & Perception & Grounding \\
\midrule

\multirow{2}{*}{50K}
& Caption & 47.25 & 42.60 & 37.86 \\
& MMCS    & 49.68 & 49.12 & 47.38 \\

\midrule

\multirow{2}{*}{600K}
& Caption & 48.95 & 49.16 & 46.61 \\
& MMCS    & 50.47 & 51.21 & 58.29 \\

\midrule

\multirow{2}{*}{1M}
& Caption & 48.60 & 48.78 & 49.85 \\
& MMCS    & 50.04 & 51.51 & 60.27 \\

\bottomrule
\end{tabular}
\caption{Performance comparison across \textbf{varying pretraining dataset sizes}. MMCS maintains an advantage over the standard baseline at 1M pretraining scales.}
\label{tab:pretrain_scale}
\end{table}

\subsection{Scaling to More Pretraining Data}
To investigate the scalability of our approach, we further expand the pretraining dataset to 1M samples by incorporating additional images from Objects365 and SA-1B. The SFT dataset is fixed to 200K for a fair comparison. All newly added samples are processed using the same data synthesis pipeline detailed in Section \ref{sec:data}. As presented in Table \ref{tab:pretrain_scale}, MMCS maintains an advantage over standard image-level pretraining across all data scales. This confirms that our explicit object-level alignment paradigm can effectively leverage larger volumes of pretraining data to achieve more robust semantic grounding.

\subsection{Statistical Significance of Improvements}
\label{app:statistical_significance}
While MMCS yields gains across all three task categories, the magnitude varies: improvements are most pronounced on grounding (+7.9\%) and perception (+2.1\%), and more modest on general VQA. To verify that the smaller gains on general VQA are statistically robust rather than artifacts of random initialization, we conduct four independent training runs under the Qwen2.5-3B + SigLIP2 configuration with different random seeds. The remaining setup is identical to that described in Section \ref{sec:setup}.

As shown in Table \ref{tab:statistical_significance}, MMCS outperforms the image-level baseline in every run, with an average improvement of +1.67\% on general VQA. A one-sided paired t-test over the four matched runs yields $p = 0.028$, indicating that this improvement is statistically significant despite its smaller magnitude. We attribute the differential gain magnitudes to the nature of each task category. MMCS directly strengthens object-level vision-language alignment, which most immediately benefits fine-grained perception and grounding. These enhanced visual capabilities also propagate to general VQA, but general VQA performance additionally depends on factors that MMCS does not directly target, such as world knowledge, commonsense reasoning, and instruction-tuning priors. As a result, MMCS produces the largest gains on perception and grounding tasks, while delivering smaller but statistically significant improvements on general VQA.

\begin{table}[t]
    \centering
    \small
    \setlength{\tabcolsep}{6pt}
    \renewcommand{\arraystretch}{1.05}
    \begin{tabular}{llccc}
    \toprule
    Method & Run & General & Perception & Grounding \\
    \midrule
    \multirow{5}{*}{Caption}
        & 1   & 51.19 & 55.77 & 59.12 \\
        & 2   & 50.49 & 54.44 & 55.59 \\
        & 3   & 50.34 & 54.38 & 62.58 \\
        & 4   & 49.13 & 54.20 & 60.55 \\
    \cmidrule(lr){2-5}
        & AVG & 50.29 & 54.69 & 59.46 \\
    \midrule
    \multirow{5}{*}{MMCS}
        & 1   & 51.84 & 58.24 & 70.29 \\
        & 2   & 51.35 & 56.99 & 70.45 \\
        & 3   & 52.59 & 56.85 & 67.17 \\
        & 4   & 52.05 & 58.02 & 71.30 \\
    \cmidrule(lr){2-5}
        & AVG & 51.96 & 57.52 & 69.80 \\
    \bottomrule
    \end{tabular}
    \caption{Multi-seed comparison between image-level captioning and MMCS under the Qwen2.5-3B + SigLIP2 setting. Each block reports four independent training runs with different random seeds. AVG denotes the per-column mean.}
    \label{tab:statistical_significance}
\end{table}

\section{Representation Alignment Metrics}
\label{app:alignment_metrics}

In this section, we provide detailed definitions of the representation alignment metrics used in Section \ref{sec:representation_alignment}. We adopt the notation from \citet{platonic}.

\paragraph{CKA} CKA (Centered Kernel Alignment) reflects the \textit{global} similarity between two representation spaces by comparing their kernel matrices. Let $\phi_i \in \mathbb{R}^n$ and $\psi_i \in \mathbb{R}^m$ be vectorized features of two modalities (e.g. language and vision). Let $\mathbf{K}_{ij} = \kappa(\phi_i, \phi_j)$ and $\mathbf{L}_{ij} = \kappa(\psi_i, \psi_j)$ be the kernel matrices computed from a dataset using some kernel function $\kappa$. For an inner-product kernel, the $ij$-th entry of the centered counterpart of these kernel matrices is given by
\begin{equation}
\begin{split}
    \bar{\mathbf{K}}_{ij} = \langle \phi_i, \phi_j \rangle - \mathbb{E}_l[\langle \phi_i, \phi_l \rangle], \\
    \bar{\mathbf{L}}_{ij} = \langle \psi_i, \psi_j \rangle - \mathbb{E}_l[\langle \psi_i, \psi_l \rangle].
\end{split}
\end{equation}
Then the cross-covariance of $\mathbf{K}$ and $\mathbf{L}$ is:
\begin{equation}
    \text{HSIC}(\mathbf{K}, \mathbf{L}) = \frac{1}{(n-1)^2} \text{Trace}(\bar{\mathbf{K}} \bar{\mathbf{L}}).
\label{eq:HSIC}
\end{equation}
Finally, CKA is obtained by normalizing this quantity:
\begin{equation}
    \text{CKA}(\mathbf{K}, \mathbf{L}) = \frac{\text{HSIC}(\mathbf{K}, \mathbf{L})}{\sqrt{\text{HSIC}(\mathbf{K}, \mathbf{K}) \text{HSIC}(\mathbf{L}, \mathbf{L})}}.
\end{equation}

\paragraph{CKNNA} CKNNA (Centered Kernel Nearest-Neighbor Alignment) is a relaxed variation of CKA that emphasizes \textit{local} structural alignment. It modifies the measure by replacing $\text{HSIC}(\mathbf{K}, \mathbf{L})$ with $\text{Align}(\mathbf{K}, \mathbf{L})$, which computes Eq. \ref{eq:HSIC} considering only the $k$-nearest neighbors in the dataset:
\begin{equation}
    \text{Align}(\mathbf{K}, \mathbf{L}) = \sum_i \sum_j \alpha(i,j) \bar{\mathbf{K}}_{ij} \bar{\mathbf{L}}_{ij},
\end{equation}
where $\alpha(i,j)$ is an indicator function that selects common nearest neighbors:
\begin{equation}
    \alpha(i,j) = \mathbbm{1} [\phi_j \in \text{knn}(\phi_i) \wedge \psi_j \in \text{knn}(\psi_i) \wedge i \neq j].
\end{equation}
This term acts as a mask, preserving only the interactions between sample $i$ and $j$ if $j$ is a neighbor of $i$ in both representation spaces.
The normalized CKNNA score is then defined as:
\begin{equation}
    \text{CKNNA}(\mathbf{K}, \mathbf{L}) =
    \frac{\text{Align}(\mathbf{K}, \mathbf{L})}
    {\sqrt{\text{Align}(\mathbf{K}, \mathbf{K})\text{Align}(\mathbf{L}, \mathbf{L})}}.
\end{equation}

\paragraph{Mutual k-NN} Mutual k-NN (Mutual k-Nearest Neighbor) measures the average overlap of nearest neighbor sets of representations. Let $\{ \phi_i, \psi_i \}_{i=1}^b$ denote a mini-batch of paired features from two modalities, where the collections of these features are denoted as $\Phi = \{ \phi_1, \dots, \phi_b \}$ and $\Psi = \{ \psi_1, \dots, \psi_b \}$. For each feature pair $(\phi_i, \psi_i)$, we compute the respective nearest neighbor sets $\mathcal{S}(\phi_i)$ and $\mathcal{S}(\psi_i)$:
\begin{equation}
\begin{split}
    \mathcal{S}(\phi_i) = d_\text{knn}(\phi_i, \Phi \setminus \phi_i ), \\
    \mathcal{S}(\psi_i) = d_\text{knn}(\psi_i, \Psi \setminus \psi_i ),  
\end{split}
\end{equation}
where $d_\text{knn}$ returns the set of indices of the $k$-nearest neighbors. We then measure the alignment via the average intersection:
\begin{equation}
    m_\text{NN}(\phi_i, \psi_i) = \frac{1}{k} |\mathcal{S}(\phi_i) \cap \mathcal{S}(\psi_i)|,
\end{equation}
where $| \cdot |$ denotes the cardinality of the set.

In Section \ref{sec:representation_alignment}, CKA is calculated using all 1000 image-caption pairs in the dataset. For CKNNA and Mutual k-NN, we report results with $k=10$.

\section{More Evaluation Results}
\label{app:more_results}

Tables \ref{tab:full_grounding}, \ref{tab:full_perception}, and \ref{tab:full_general} report the per-benchmark results that underlie the averaged scores, covering visual grounding, perception-centric, and general VQA benchmarks, respectively.

\begin{table*}[t]
    \centering
    \small
    \begin{tabular*}{0.8\textwidth}{@{\extracolsep{\fill}}lccccccccc}
    \toprule
    \multirow{2}{*}{Method}
     & \multicolumn{3}{c}{RefCOCO}
     & \multicolumn{3}{c}{RefCOCO+}
     & \multicolumn{2}{c}{RefCOCOg}
     & \multirow{2}{*}{AVG} \\
    \cmidrule(lr){2-4}\cmidrule(lr){5-7}\cmidrule(lr){8-9}
     & testA & testB & val & testA & testB & val & test & val & \\
    \midrule
    \rowcolor{gray!15}\multicolumn{10}{l}{\textit{Qwen2.5-3B + SigLIP2}}\\
    \addlinespace[2pt]
    Caption        & 65.00 & 54.64 & 61.68 & 63.88 & 47.49 & 56.62 & 61.27 & 62.39 & 59.12 \\
    SEA            & 68.33 & 60.46 & 64.08 & 62.81 & 48.02 & 57.80 & 61.09 & 65.05 & 60.96 \\
    Patch Aligned  & 77.82 & 64.08 & 71.84 & 71.34 & 50.83 & 61.95 & 66.67 & 67.90 & 66.55 \\
    Text BBox      & 75.91 & 60.35 & 67.60 & 69.37 & 49.15 & 60.30 & 65.03 & 67.83 & 64.44 \\
    MMCS           & \textbf{80.66} & \textbf{68.75} & \textbf{75.18} & \textbf{74.35} & \textbf{56.23} & \textbf{65.49} & \textbf{70.32} & \textbf{71.31} & \textbf{70.29} \\
    \quad w/o $\mathcal{L}_\text{entity}$ & 78.50 & 64.59 & 73.18 & 72.14 & 52.38 & 63.06 & 68.04 & 68.49 & 67.55 \\
    \quad w/o $\mathcal{L}_\text{LM}$     & 77.60 & 62.65 & 71.52 & 69.87 & 49.85 & 60.55 & 65.06 & 66.43 & 65.44 \\
    \bottomrule
    \end{tabular*}
    \caption{More results on referring expression comprehension. We report Acc@0.5 across all splits.}
    \label{tab:full_grounding}
\end{table*}

\begin{table*}[t]
    \centering
    \small
    \begin{tabular*}{0.8\textwidth}{@{\extracolsep{\fill}}lcccccccc}
    \toprule
    Method & AI2D & ChartQA & CVBench & OCR & RWQA & VQA$^\text{T}$ & V* & Avg \\
    \midrule
    \rowcolor{gray!15}\multicolumn{9}{l}{\textit{Qwen2.5-3B + SigLIP2}}\\
    \addlinespace[2pt]
    Caption        & 66.13 & 56.52 & 61.26 & 44.70 & 55.29 & 58.29 & 48.17 & 55.77 \\
    SEA            & 66.77 & 53.48 & 63.15 & 44.90 & 55.42 & 58.48 & 47.64 & 55.69 \\
    Patch Aligned  & 67.45 & 56.08 & 61.87 & 46.70 & 55.16 & 59.58 & 48.17 & 56.43 \\
    Text BBox      & 67.10 & 56.60 & 65.39 & 44.00 & 54.64 & 58.38 & 50.26 & 56.62 \\
    MMCS           & \textbf{68.72} & 57.36 & \textbf{66.87} & \textbf{47.60} & \textbf{56.99} & 59.35 & \textbf{50.79} & \textbf{58.24} \\
    \quad w/o $\mathcal{L}_\text{entity}$ & 67.10 & 54.40 & 60.88 & 43.80 & 52.94 & 57.57 & 42.93 & 54.23 \\
    \quad w/o $\mathcal{L}_\text{LM}$     & 67.13 & \textbf{57.40} & 61.60 & 46.30 & 53.72 & \textbf{60.04} & 47.64 & 56.26 \\
    \midrule
    \rowcolor{gray!15}\multicolumn{9}{l}{\textit{Qwen2.5-3B + QwenViT}}\\
    \addlinespace[2pt]
    Caption & 69.40 & 70.36 & 56.33 & 58.80 & 55.16 & 62.87 & 57.59 & 61.50 \\
    MMCS    & \textbf{71.15} & \textbf{71.92} & \textbf{59.78} & \textbf{62.40} & \textbf{56.86} & \textbf{64.97} & 57.59 & \textbf{63.52} \\
    \midrule
    \rowcolor{gray!15}\multicolumn{9}{l}{\textit{Qwen3-8B + SigLIP2}}\\
    \addlinespace[2pt]
    Caption & 72.51 & 62.48 & 69.71 & 50.60 & 58.17 & 64.37 & 50.79 & 61.23 \\
    MMCS    & \textbf{73.93} & \textbf{64.96} & \textbf{72.82} & \textbf{53.30} & \textbf{60.13} & \textbf{65.23} & \textbf{53.40} & \textbf{63.40} \\
    \midrule
    \rowcolor{gray!15}\multicolumn{9}{l}{\textit{Llama3-8B + SigLIP2}}\\
    \addlinespace[2pt]
    Caption & \textbf{72.31} & 58.28 & 65.66 & 47.00 & \textbf{59.87} & 62.59 & 51.31 & 59.57 \\
    MMCS    & 71.96 & \textbf{60.20} & \textbf{69.48} & \textbf{48.90} & 59.08 & \textbf{64.50} & \textbf{53.93} & \textbf{61.15} \\
    \bottomrule
    \end{tabular*}
    \caption{More results on perception-centric benchmarks. ``OCR'', ``RWQA'', ``VQA$^\text{T}$'', and ``V*'' denote OCRBench, RealWorldQA, TextVQA, and V-Star, respectively.}
    \label{tab:full_perception}
\end{table*}

\begin{table*}[t]
    \centering
    \small
    \begin{tabular*}{0.8\textwidth}{@{\extracolsep{\fill}}lccccccc}
    \toprule
    Method & MMB & MME & MMMU & MMStar & MMVet & GQA & Avg \\
    \midrule
    \rowcolor{gray!15}\multicolumn{8}{l}{\textit{Qwen2.5-3B + SigLIP2}}\\
    \addlinespace[2pt]
    Caption        & 67.53 & 60.36 & 40.60 & 42.77 & 35.64 & 60.22 & 51.19 \\
    SEA            & 68.81 & 60.86 & 40.40 & 42.54 & 31.83 & 60.26 & 50.78 \\
    Patch Aligned  & \textbf{69.76} & \textbf{63.71} & \textbf{41.30} & 43.45 & 32.11 & 60.36 & 51.78 \\
    Text BBox      & 67.87 & 60.32 & 41.10 & \textbf{44.41} & \textbf{35.69} & 60.45 & 51.64 \\
    MMCS           & 68.56 & 62.75 & 40.90 & 43.98 & 33.30 & \textbf{61.57} & \textbf{51.84} \\
    \quad w/o $\mathcal{L}_\text{entity}$ & 66.75 & 63.50 & 40.80 & 44.00 & 33.02 & 60.61 & 51.45 \\
    \quad w/o $\mathcal{L}_\text{LM}$     & 69.07 & 61.82 & 40.20 & 43.71 & 33.03 & 60.41 & 51.37 \\
    \midrule
    \rowcolor{gray!15}\multicolumn{8}{l}{\textit{Qwen2.5-3B + QwenViT}}\\
    \addlinespace[2pt]
    Caption & 67.70 & 62.82 & 39.20 & 45.21 & 39.27 & 60.12 & 52.39 \\
    MMCS    & \textbf{68.47} & \textbf{65.68} & \textbf{42.90} & \textbf{45.31} & \textbf{39.31} & \textbf{60.66} & \textbf{53.72} \\
    \midrule
    \rowcolor{gray!15}\multicolumn{8}{l}{\textit{Qwen3-8B + SigLIP2}}\\
    \addlinespace[2pt]
    Caption & 74.48 & 68.14 & 47.10 & 48.67 & \textbf{44.72} & 62.54 & 57.61 \\
    MMCS    & \textbf{76.63} & \textbf{68.54} & \textbf{48.80} & \textbf{51.49} & 41.56 & \textbf{63.02} & \textbf{58.34} \\
    \midrule
    \rowcolor{gray!15}\multicolumn{8}{l}{\textit{Llama3-8B + SigLIP2}}\\
    \addlinespace[2pt]
    Caption & \textbf{70.79} & 63.43 & 39.20 & \textbf{44.96} & 39.36 & 63.17 & 53.48 \\
    MMCS    & 68.99 & \textbf{65.54} & \textbf{40.90} & 43.15 & \textbf{42.06} & \textbf{63.40} & \textbf{54.01} \\
    \bottomrule
    \end{tabular*}
    \caption{More results on general VQA benchmarks. We report the normalized score for MME, and evaluate MMVet using GPT-4.1 as the judge.}
    \label{tab:full_general}
\end{table*}

\section{More Qualitative Results}
\label{app:more_attn_results}

\begin{figure*}[t]
    \centering
    \includegraphics[width=\textwidth]{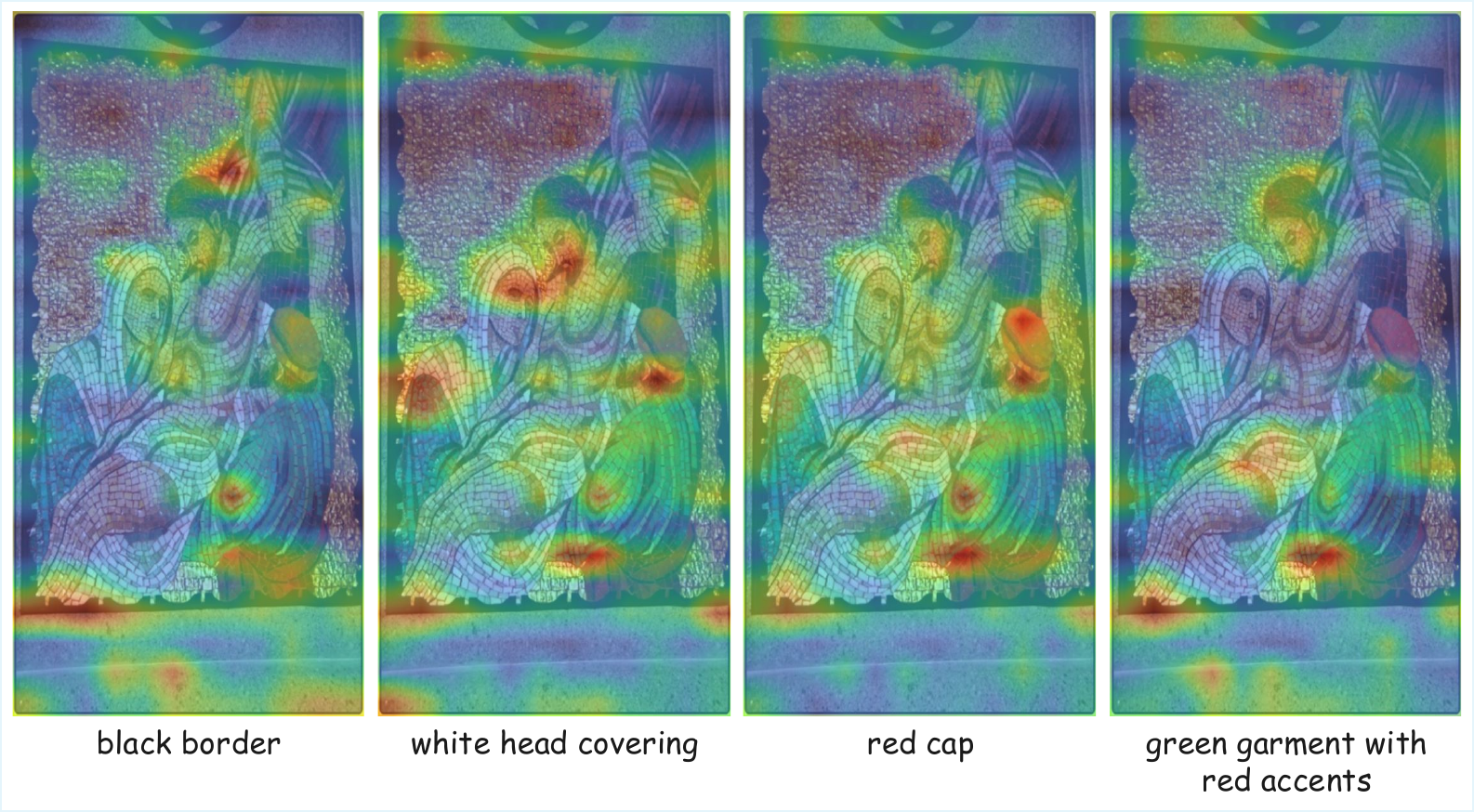}
    \caption{Visualization of dynamic attention shifts for a complex scene containing multiple target objects. The heatmaps illustrate how the model's visual attention shifts to focus on the specific region as the corresponding textual entity is processed.}
    \label{fig:attn_vis_multiple}
\end{figure*}

Figure \ref{fig:attn_vis_multiple} presents a representative case demonstrating our model's dynamic attention behavior when processing scenes with multiple target objects. Specifically, given an input image of a framed mosaic artwork alongside a dense caption, we observe clear spatial shifts in the model's visual attention as it processes different textual entities. This capability indicates that our model can effectively disentangle multiple localized objects within a single complex image. Furthermore, this dynamic localization behavior suggests that the explicit object-entity correspondence established by our MMCS paradigm successfully guides the model to decouple semantic regions.

\end{document}